\documentclass[12pt]{article}
\usepackage{amsmath}
\usepackage{graphicx}
\usepackage{enumerate}
\usepackage{natbib}
\usepackage{url}
\usepackage{color}
\usepackage{float}
\usepackage[utf8]{inputenc}
\usepackage{amsthm} 
\usepackage{amssymb}
\usepackage{multirow}
\usepackage{latexsym}
\usepackage{enumitem}

\newcommand{\blind}{1}

\usepackage{float}
 
\begin{document}
	
	
	\def\spacingset#1{\renewcommand{\baselinestretch}%
		{#1}\small\normalsize} \spacingset{1}

	
	\if1\blind
	{
		\title{\bf Towards the Best Solution for Complex System Reliability: Can Statistics Outperform Machine Learning? 
		}
		\author{
			Mar{\'\i}a Luz G\'amiz \\ Department of Statistics and O.R., University of Granada, Spain\\ \\
		Fernando Navas-G\'omez\\  Department of Statistics and O.R., University of Granada, Spain\\ \\
		Rafael Nozal-Ca\~nadas	\\ Department of Computer Science, UiT-The Arctic University of Norway, Norway\\ \\
		Roc\'io Raya-Miranda \thanks{Corresponding author: rraya@ugr.es}\\
		 Department of Statistics and O.R., University of Granada, Spain
		}
		
		\maketitle
	} \fi
	
	\if0\blind
	{
		\bigskip
		\bigskip
		\bigskip
		\begin{center}
			{\LARGE\bf Towards the Best Solution for Complex System Reliability: Can Statistics Outperform Machine Learning?}
		\end{center}
		\medskip
	} \fi
	
	\bigskip
	
	\begin{abstract}
		{Studying the reliability of complex systems using machine learning techniques involves facing a series of technical and practical challenges, ranging from the intrinsic nature of the system and data to the difficulties in modeling and effectively deploying models in real-world scenarios. This study compares the effectiveness of classical statistical techniques and machine learning methods for improving complex system analysis in reliability assessments. We aim to demonstrate that classical statistical algorithms often yield more precise and interpretable results than black-box machine learning approaches in many practical applications. The evaluation is conducted using both real-world data and simulated scenarios. We report the results obtained from statistical modeling algorithms, as well as from machine learning methods including neural networks, K-nearest neighbors, and random forests.
}
\end{abstract}

\noindent%
{\it Keywords:}Logistic Regression; Factorial Analysis; Isotonic Smoothing; Machine Learning;  Supervised Learning; Unsupervised Learning; ANN; KNN; RF
\vfill

\newpage
\spacingset{1.5} 

\section{Introduction}
\label{sec:intro}

Reliability analysis of complex, multicomponent systems is crucial in engineering, manufacturing, and operations research. It involves understanding and quantifying the capacity of a system to perform its intended function over a specified period under given conditions. Addressing the reliability of such systems requires a multifaceted approach that integrates structural analysis, probabilistic modeling, and practical maintenance strategies. These methodologies help engineers design more reliable systems, predict potential failures, and develop effective maintenance plans to mitigate risks.

However, as the number of components increases, the lack of knowledge about the structure of the system leads to an estimation problem with an overwhelming number of features \cite{Gamizetal2023}. This results in exponential growth in dimensionality and sparse available data. Consequently, high-dimensional reliability analysis remains a significant challenge, as most existing methods suffer from the so called {\em curse of dimensionality}  \cite{LiWang20}.

Machine learning (ML), a form of applied statistics, focuses on using computational power to estimate complex functions, unlike traditional statistics, which prioritize confidence intervals and rigorous proofs of uncertainty \cite{GoodfellowBC2016}.  ML employs algorithms to detect patterns in data, predicting and assessing performance and failure risks in systems. By doing so, it enhances design, maintenance, and safety. ML is defined as a set of methods that uncover patterns in data to predict future outcomes or support decision-making under uncertainty \cite{Murphy12}. Over recent decades, ML has revolutionized fields such as control systems, autonomy, and computer vision. Similarly, reliability engineering and safety analysis are expected to follow this trend  \cite{XuSaleh21}.

ML offers several advantages over traditional methods, including advanced predictive capabilities, the ability to handle large, high-dimensional datasets, real-time monitoring, and automation, all of which contribute to improved system reliability. Common applications of ML in reliability engineering include estimating remaining useful life, anomaly and fault detection, health monitoring, maintenance planning, and degradation assessment. Studies like  \cite{XuSaleh21} highlight the potential of ML to transform reliability and safety analysis. However, challenges such as data quality and model interpretability must be addressed for ML to fully realize its potential (see \cite{Gamizetal2023} and references therein). Interdisciplinary collaboration and hybrid approaches also show promise \cite{Gupta2022, Sudiana2023, Gamizetal2023}. For instance, \cite{Gupta2022} proposes a hybrid model combining logistic regression, multilayer perceptron, and XGBoost, while  \cite{Sudiana2023} combines a convolutional neural network with a random forest for burned area detection. Similarly,  \cite{Gamizetal2023} integrates supervised and unsupervised learning techniques to analyze system status, and \cite{Daya24} combines reliability analysis tools with ML to identify critical maintenance components and failure causes.

When applying ML, the question of which model performs best often arises. In 1997, David Wolpert and William MacReady addressed this issue with the "No Free Lunch" (NFL) theorem \cite{Wolpert1997}, which demonstrated that without assumptions about the data, no model is universally superior. This implies that no model can consistently outperform others in all scenarios. As a result, selecting the best model requires evaluating multiple options, which is often impractical. In practice, assumptions about the data lead researchers to focus on a subset of models deemed appropriate  \cite{Adam19}.

The use of ML for reliability analysis in large-scale, complex systems is a key application within Industry 4.0. In this context, the interconnection of sensors and smart devices generates large volumes of data, which can be analyzed to enhance system efficiency and safety. In systems with thousands of sensors, traditional reliability models based on statistical distributions become inefficient. Machine learning techniques, such as deep learning, are better suited for handling high-dimensional data. Algorithms like Random Forest and Bayesian classification models can extract valuable information from the vast datasets produced by sensors \cite{Peng2019}. Industry 4.0 now includes diverse data sources, such as temperature, vibration, and pressure sensors. The application of ML algorithms enables the integration and simultaneous analysis of these variables, detecting correlations that might not be apparent with traditional techniques \cite{DiNoia2021}.

The paper is structured as follows: Section \ref{Sec:Ml} describes methods for estimating reliability in complex, high-dimensional systems, and discusses machine learning solutions to this problem, along with metrics for evaluating classification methods. Section \ref{Sec:FA} introduces the FA-LR-IS algorithm. Section \ref{Sec:results} presents numerical results from both a simulation study and an application to a real dataset. Section \ref{Sec:disc} provides a brief discussion on the topic, and Section  \ref{Sec:conc} concludes with the conclusions.

\section{Some ML methodologies}\label{Sec:Ml}
Reliability analysis is fundamental in the design and maintenance of complex systems. One key question is that, traditionally, systems have been analysed under the assumption that their components are independent, but this assumption becomes less valid as systems grow in complexity.  In reality, system components are often interdependent, they are designed to work together to ensure the proper operation of the system. Addressing these interdependencies is crucial for improving the accuracy of reliability predictions. A complex system is characterised by the nonlinear interaction of its components, where the dynamics of a system cannot be understood simply as the sum of its parts. These systems exhibit emergent properties, and adaptability, and may have hierarchical structures. To model such complex interactions, advanced methods like ML can be employed, which allow for the consideration of nonlinear relationships between components.

The analysis of complex systems in the context of reliability needs to address these interdependencies to improve the accuracy of reliability predictions. To accurately assess the reliability of complex systems, traditional methods that assume independent components are insufficient. This is where machine learning (ML) methods, such as neural networks, become valuable. These methods can model the complex interdependencies between the components of the system, providing more accurate predictions. This capability makes them ideal for reliability prediction, especially in systems with numerous components that interact in ways that traditional methods may not easily capture. Neural networks, in particular, offer significant potential in this context, having demonstrated success in various fields such as image recognition, speech recognition, and natural language processing. However, until 2006 we did not know how to train neural networks to surpass more traditional approaches \cite{Nielsen2015}, except for a few specialised problems. In a neural network, we do not tell the computer how to solve our problem, instead, it learns from observational data, figuring out its solution to the problem at hand. What changed in 2006 was the discovery of learning techniques in so-called deep neural networks \cite{Nielsen2015}.

ML methods have to be adapted according to the area of application and it is necessary to incorporate the knowledge of experts to improve the accuracy and reliability of the predictions \cite{Hussain24}. Below are some of the ML methods that can be applied to predict the reliability of a complex system with multiple components and possible correlations between them.

To formalize the reliability prediction problem, let us define a set of random variables representing the components of the system and its operational status. Let $\textbf{X}=(X_1,\ldots,X_p)$ where $X_k$ is a random variable representing the level of performance of the $k$-th component of the system. We assume that $X_k \in [0,1]$, $k=1,\ldots,p$ and let $Y$ be a binary random variable taking value 1 where the system is working and 0 otherwise. 

Let $\{\textbf{X},Y\}$ be the observed data where $\textbf{X}$ is a matrix of dimension $n \times p$. Each row of the matrix is a configuration of the components states of system $i$, with $i=1,2,\ldots,n$. $Y$ is the $n$-dimensional vector of all systems of the sample. The objective is to predict whether the system is operational given the state of the components. In the following, we describe the basic elements of some ML methods that we will apply in our numerical analysis in Section \ref{Sec:results}.

\subsection{Artificial Neural Networks}
Artificial neural networks (ANN) are ML algorithms that simulate the learning mechanisms of biological organisms \cite{Aggarwal2018}. The application of neural networks in reliability estimation is based on their ability to model complex relationships between input data (e.g. historical failure data or operating conditions) and output results (such as failure probability, remaining useful life, etc.). Neural networks significantly improve the ability to anticipate failures, optimize maintenance, and ensure reliability in complex systems, especially in industries where the cost of a failure can be high  \cite{Li2018}.

ANNs are made up of units called neurons, which are interconnected in a structure consisting of at least two layers: an input layer and an output layer. In the case of having a hidden layer, it must have at least one hidden layer. The input layer receives the data (relevant features such as operating conditions, runtime, sensor variables, etc.), while the hidden layers process the information in an intermediate manner. Each neuron in these layers performs a mathematical transformation based on the learned weights and biases. Finally, the output layer produces the reliability estimate, such as the probability of failure or the remaining lifetime. Each neuron takes a linear combination of the inputs and then applies a nonlinear activation function. This process can be described as follows:
$$
z_j^{(l)} = \sum_{i=1}^n w_{ij}^{(l)} x_i^{(l-1)} + b_j^{(l)}
$$
where:
\begin{itemize}
	\item $z_j^{(l)}$ is the value of neuron $j$ in layer $l$.
	\item $w_{ij}^{(l)}$ is the weight connecting neuron $i$ in the previous layer $l-1$ to neuron $j$ in the current layer $l$.
	\item $x_i^{(l-1)}$ is the output of neuron $i$ in the previous layer.
	\item $ b_j^{(l)}$ is the bias of neuron $j$ in layer $l$.
	\item $n$ is the number of neurons in the previous layer.
\end{itemize}
Then, an activation function $g$ is applied to introduce nonlinearity, $ a_j^{(l)}=g(z_j^{(l)})
$; for more details \cite{Aggarwal2018, Anthony2010}. Depending on the type of study, selecting the appropriate activation function allows the network to learn more complex patterns and perform more sophisticated tasks. For reliability classification problems, activation functions such as ReLU (Rectified Linear Unit) or the sigmoid are commonly used:
\begin{itemize}
	\item ReLU: $g(z) = \max (0,z)$
	\item Sigmoid: $g(z)= \frac{1}{1+e^{-z}} $
\end{itemize}

The sigmoid function is especially popular in the output layer for binary classification problems, as it generates an output in the form of a probability \cite{Singh23}. This process is repeated in all the hidden layers. In the last layer (the output layer), the output value $Y$, which represents the probability of failure at a given time, is generated using an activation function such as the sigmoid in the case of classification problems.

The ANN requires a learning process to adjust the weights and biases of the connections, which incurs a computational cost. Algorithms such as gradient descent (GD), stochastic gradient descent (SGD), adaptive gradient descent (AdaGrad), and root mean square propagation (RMSprop) are commonly used to minimize the loss function \cite{Aggarwal2018, Duchi2011, Kingma2017}. 

Loss functions are critical in the training and validation stages, as they minimize the difference between the system state predicted by the model and the actual state, allowing for correct fit of the model parameters, i.e., weights and biases. For regression problems, the mean square error is typically used, while binary cross-entropy is an appropriate cost function for binary classification problems.

\subsection{K-Nearest Neighbors}
The K-Nearest Neighbors (KNN) algorithm is a supervised learning technique commonly used in classification and regression problems. In the context of reliability, KNN can be applied to make predictions related to the probability of failure or lifetime of a system based on historical data of similar failures. An example would be the prediction of the remaining lifetime of electronic devices under accelerated stress conditions, where machine learning models such as KNN are used to estimate the reliability of electronic components with high levels of accuracy \cite{Qiu2024}.

Using KNN, a system is classified based on data from other similar systems (neighbors). If a majority of the nearby neighbors have failed under similar conditions, the algorithm predicts that the system is also at risk of failure. The aim is to assign an unclassified point ${\bf x}$, i.e., a given configuration of component states, to the class state of the system represented by a majority of its $K$-nearest neighbors. To do this, the distance between the point of interest ${\bf x}$ and each point ${\bf x}_{i}$ in the data set, $d({\bf x},{\bf x}_{i})$ is calculated. The $K$ nearest neighbors are selected, that is, those points that minimize the distance $d({\bf x},{\bf x}_{i})$. Finally, the majority class of the state of the system is determined among the nearest neighbors; the system state,  $\hat{Y}\in Y$, for point ${\bf x}$ will be the one that appears most frequently among the $K$ neighbors.

The choice of the parameter $K$ has a significant influence on the performance of the model \cite{Zhan2006, Ghosh2006, Li2019}. The importance of normalizing the data is also highlighted since the KNN algorithm is sensitive to the scale of the variables.  This algorithm is simple to implement and easy to understand \cite{Guo2003}, but it can have a high computational cost if working with large datasets due to the calculation of distances. 

To choose the parameter $K$ correctly, there are several alternatives, such as opting for an odd value of $K$, which would avoid possible ties in the proportions of membership in each class. The leave-one-out or $K$-fold cross-validation technique consists of dividing the data set into $K$ subsets, using $K-1$ of these to train the model and the remaining subset to evaluate performance. This process is repeated $K$ times and the model performance is averaged for different values of $K$. Another option is to try different values of $K$, apply the method to sample points whose classification is known and select the value of $K$ that minimizes the classification error. 

Empirically, one can highlight the square root rule of the data set size, which suggests that $K$ should be approximately equal to the square root of the total size of the dataset. In KNN algorithms, the choice of the distance metric directly affects how the proximity between points in the feature space is calculated, which in turn influences classification decisions. The Euclidean distance is the most common metric; however, the Manhattan distance can also be used. Finally, we can mention the Minkowski distance, which generalizes the two previous ones and whose expression is given by 
$$d({\bf x},{\bf x}_{i})=\left(\sum_{i=1}^{n}|{\bf x}-{\bf x}_{i}|^{q}\right)^{1/q}.$$
When $q=2$, it is equivalent to the Euclidean distance, and when $q=1$, it is equivalent to the Manhattan distance.

\subsection{Random Forest}
Random Forest (RF) is a powerful ML technique that has been applied in the field of reliability engineering to improve failure prediction, risk assessment, and maintenance decision-making. Its ability to handle complex, high-dimensional data, as well as providing insight into the most relevant factors affecting system reliability, makes it a particularly valuable tool in real-world applications. Several scientific studies have explored the use of RF in reliability engineering; for example, to predict failures in semiconductor manufacturing equipment, demonstrating the high accuracy and robustness of the algorithm in handling complex operational data \cite{Cheng2013}, and to estimate product failures and warranty costs, highlighting the effectiveness of the model in analyzing warranty claims data and in projecting future costs based on historical failure data \cite{Tiwari2017}. 

In the reliability domain, RF is based on the construction of multiple independent and uncorrelated decision trees \cite{Breiman2001}. Each tree is trained using a random portion of the training dataset and a random subset of component states at each split of the tree. This process, known as bagging or bootstrap aggregating, helps reduce model variance and prevent overfitting. Once all decision trees are trained, the RF algorithm makes predictions by combining the individual predictions from each tree. In classification problems, a majority voting strategy is used: the system state that receives the most votes among all trees is selected as the final prediction. The RF predictor is defined as the average of the predictions from $B$ independent decision trees, each trained on a randomly selected dataset:
$$\hat{Y} = \frac{1}{B}\sum_{i=1}^B T_b({\bf x}_{i}),
$$
where $B$ is the total number of trees in the forest, and $T_b$ is the $b$-th decision tree. In the computational implementation, each tree is built following a detailed procedure:
\begin{enumerate}
	\item Subset selection: Training samples of component states are randomly selected, with replacement, for each tree.
	\item Node splitting: Nodes in each tree are split using the best possible partition within a random subset of component states, optimizing a purity criterion. The most common measure for classification is the Gini impurity or information gain (based on entropy)
	$$
	G(p) = 1 - \sum_{k=1}^K  p^2_k,
	$$
	where $p_k$ is the proportion of examples at the node that belongs to class $k$, and $K$ is the total number of classes. Splitting of each node is done in such a way as to minimize impurity.
	
	\item Tree growth: Trees are expanded to the maximum allowed size, without pruning except for explicit constraints such as maximum depth or minimum number of samples per leaf.
\end{enumerate}

The component state importance is based on the sum of the Gini impurity reduction, weighted by the number of samples arriving at each node, and averaged across all trees.

This approach makes RF particularly effective at handling large data sets with numerous features, providing efficient computational performance tailored to solving complex problems in ML \cite{Zhao2019}. 

\subsection{Evaluating the methods}\label{sec:metrics}
\begin{figure}[ht!]
		\centering
		\includegraphics[width=0.6\textwidth]{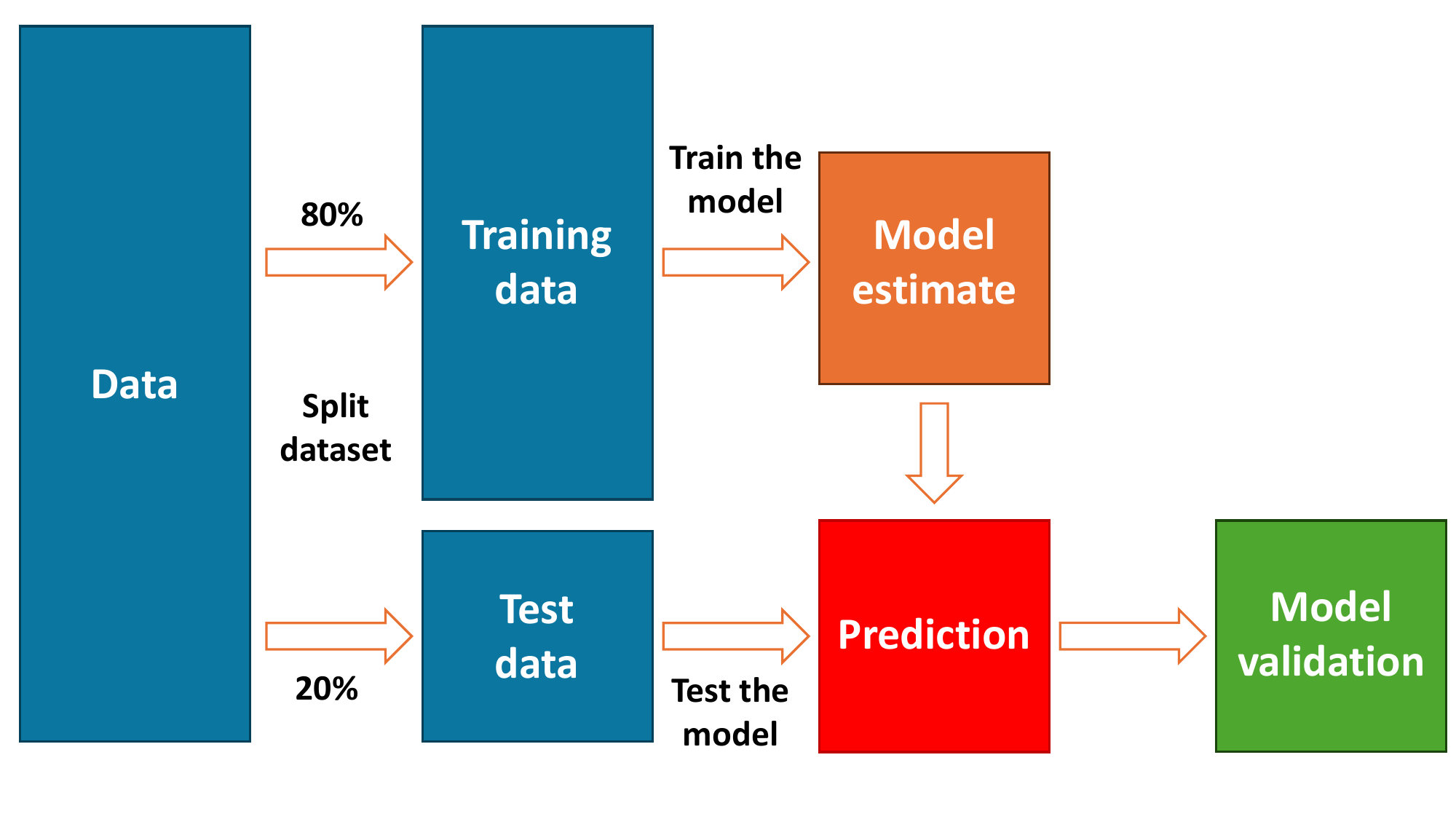}
	\caption{Splitting datasets.}\label{fig:tt}
\end{figure}

To evaluate and compare the proposed algorithm that combines unsupervised and supervised learning methods with other supervised methods, the algorithm is presented with the classic procedure of division into training and test data, which is common in this context, as shown in Figure \ref{fig:tt}.

\begin{figure}[ht!]
		\centering
		\includegraphics[width=0.6\textwidth]{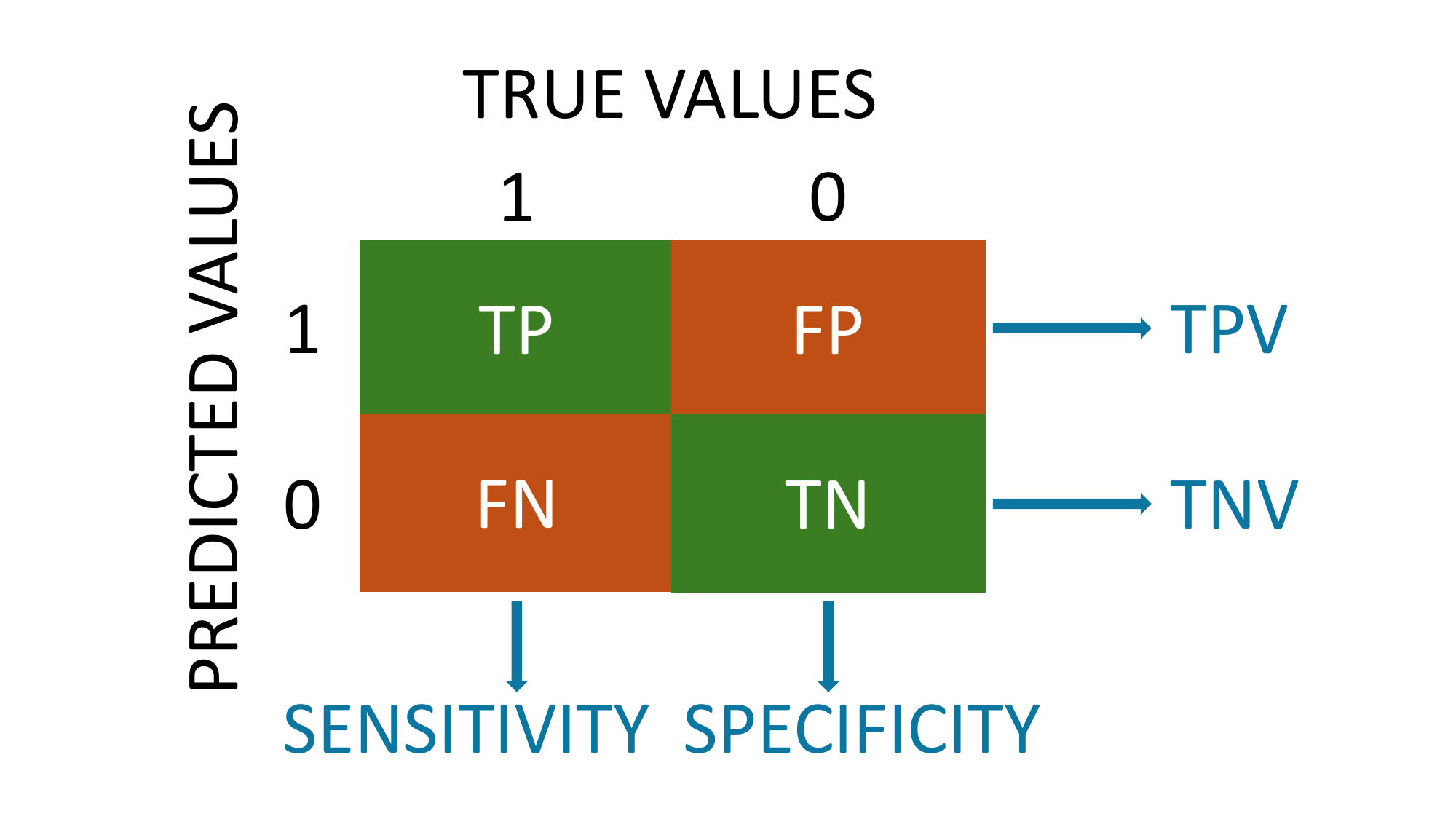}
	\caption{Confusion matrix.\label{fig:ConfM}}
\end{figure}

The most common error metrics in the context of ML algorithms for classification, as it is our case, are based on a confusion matrix based on the values of the original response variable and the values predicted by the algorithm. The matrix has four entries that we call true positive (TP), true negative (TN), false positive (FP), and false negative (FN), respectively, see Figure \ref{fig:ConfM}. In our context we call a "positive" when the system is working, i.e. $Y=1$, and a "negative" means that the system is in failure $Y=0$. 
Based on these measurements the error metrics can be calculated as follows:
\begin{itemize}
	\item Sensitivity: The systems are correctly classified as operative.
	$$ \text{Sensitivity}=\dfrac{\text{TP}}{\text{TP+FN}}$$
	It measures the probability with which the model correctly predicts that the system works.
	\item Specificity: The systems correctly classified as failed.
	$${\rm Specificity} =\dfrac{\text{TN}}{\text{TN+FP}}$$
	It measures the capacity (probability) of the model to detect failures in the system.
	\item Accuracy: The systems are correctly classified.
	$$
	{\rm Accuracy}=\dfrac{\text{{TP+TN}}}{\text{TP+FP+TN+FN}}
	$$
	The probability that, given a configuration of components, the model correctly classifies the system.
	\item True Positive Value: The systems are correctly classified among all systems classified as operative.
	$${\rm TPV}=\dfrac{\text{TP}}{\text{TP+FP}}$$
	The probability that the system is working when the model predicts it to be.
	\item F1-Score: Harmonic mean between TPV and sensitivity. The two metrics contribute equally to the score, ensuring that the F1 metric correctly indicates the reliability of a model.
	$${\rm F1-Score}=2 \cdot \dfrac{\text{TPV}\cdot \text{Sensitivity}}{\text{TPV} + \text{Sensitivity}}$$
	A high F1-Score generally indicates a well-balanced performance, demonstrating that the model can concurrently attain high TPV and high sensitivity.
\end{itemize}

\section{The FA-LR-IS Algorithm}\label{Sec:FA}
As we have already mentioned, one of the primary goals in reliability analysis is to mathematically represent the logic underlying a system. Assessing system performance, even for simple structures, can be challenging, making it essential to develop effective methods for modeling the relationship between the state of the system and its components. This approach helps classify complex systems. As a result, the system structure function, which links the system’s state with its components, has become central in system reliability studies.

In this section, we introduce our algorithm for building a statistical model capable of predicting system reliability for complex systems with numerous interdependent components. The goal is to estimate the likelihood of the system functioning based on its components' performance levels. However, with a large number of inputs, overfitting and higher prediction errors may arise. To address this, the algorithm first reduces dimensionality using factor analysis (FA), which groups correlated variables into factors that share common variance. This results in a reduced set of variables related to the original ones.

Next, a local logistic model is built in the latent space rather than using the original component states as inputs. The local regression model is constructed on the scores matrix generated by the FA algorithm. Nonparametric regression ensures estimators with desirable properties like consistency and normality, but these estimators are not inherently monotonic. In a coherent system, however, the regression model’s response variable (the system state) must be monotonic relative to the original variables (component states). Since the features used in the local logistic model lack clear physical interpretation, we cannot assume monotonicity. Therefore, we propose an isotonization step after back-transformation to ensure the model is expressed in terms of the original variables.

Finally, the estimated probabilities generated by the logistic regression model are translated into classes or categories using the classification obtained from the ROC curve. The area under the curve ($AUC$) is calculated as the probability that the reliability predicted for an operative system exceeds that of a failed system \cite{Gamiz2021}.  To approximate this probability, we split the sample into failed systems ($A_F$) and operative systems ($A_O$) with respective sizes $n_F$ and $n_O$, (where $n=n_F+n_O$). The linear regression model also classifies the systems of the sample in one of the two classes, 0 and 1, based on the predicted system reliability $\hat{R}(\bf{x})$. So we define the following statistic 
\begin{equation}
	\widehat{AUC} = \frac{1}{n_F\cdot n_O} \sum_{i \in A_F} \sum_{j \in A_O} I \left( \hat{R} ({\bf x}_{i}) < \hat{R} ({\bf x}_j) \right). \label{eq:AUC}
\end{equation}
The $\widehat{AUC}$ value thus be used as an assesses the classifier’s discriminatory power (its ability to distinguish between the two classes). A perfect classifier would yield an $\widehat{AUC}$=1.

\subsection{Description}\label{sec:algorithm}
Let $\{\textbf{X},Y\}$ be the observed data where ${\bf X}$ is a matrix of dimension $n \times p$. Each input into the matrix, for $i=1,\ldots, n$, represents a configuration of the components states of a system of dimension $p$. The state of each component is given by a random variable that is assumed to range in the interval $[0,1]$. The components of the system are not necessarily independent. $Y$ denotes a vector of dimension $n$. For $i=1,2,\ldots, n$, $Y_i=1$ if the system is operative, and $Y_i=0$ otherwise. Figure \ref{fig:diagramaflujo} represents the flowchart of the algorithm described below:
\begin{figure}[ht!]
	
	\centering
	\includegraphics[width=0.8\textwidth]{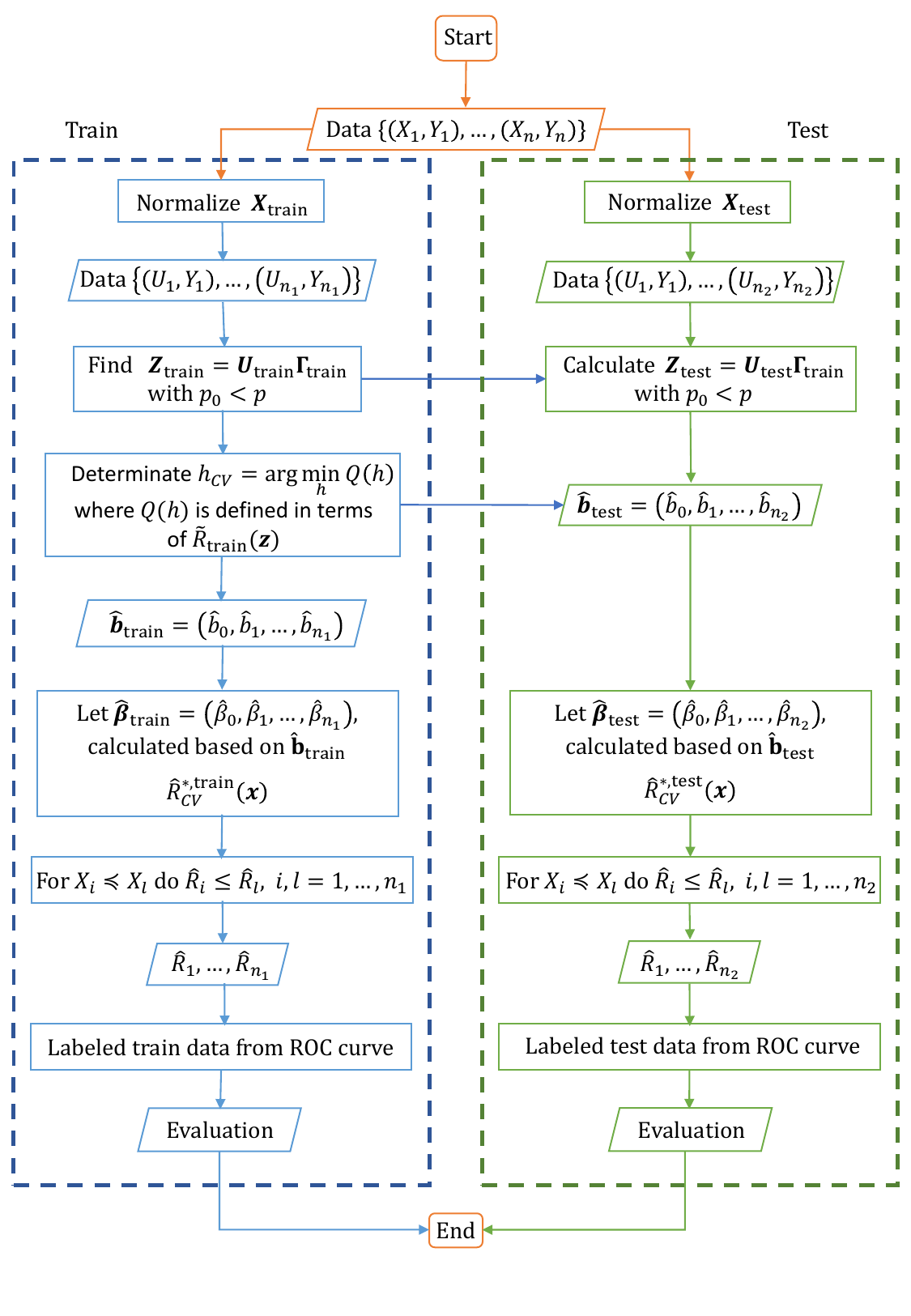}
	\caption{FA-LR-IS algorithm.}\label{fig:diagramaflujo}
	
\end{figure} 
\begin{enumerate}
	\item Split the data in two sets: train ($80\%$ data, $n_1$) and test set ($20\%$ data, $n_2$). 
	
	\item Working with train dataset:
	\begin{enumerate}
		\item Center the columns of matrix ${\bf X}_{\text{train}}$ that is, for each $j=1,2,\ldots, p$, define $U_j=(X_j-\mu_j)/\sigma_j$. Let $ {\boldsymbol \Sigma}_0$ a diagonal matrix with $j$th element $ \sigma_j$, and ${\bf M}_{\text{train}}={\bf 1}_p (\mu_1, \mu_2, \ldots, \mu_p)$, with ${\bf 1}_p$ a unitary column vector of size $p$. Let us denote ${\bf U}_{\text{train}}= \left({\bf X}_{\text{train}}- {\bf M}_{\text{train}}\right) {\boldsymbol \Sigma}_0^{-1}$, then ${\bf U}_{\text{train}}$ is a matrix with dimension $n_1 \times p$.
		
		\item Apply an algorithm for factorial analysis (FA) to the scaled dataset ${\bf U}_{\text{train}}$ and transform the data to a reduced set denoted $\{{\bf Z}_{\text{train}}, Y_{\text{train}}\}_{ n_1 \times p_0}$, with $p_0 < p$, and ${\bf Z}_{\text{train}} = {\bf U}_{\text{train}} \boldsymbol{\Gamma}_{\text{train}}$, being $\boldsymbol{\Gamma}_{\text{train}}$ the FA coefficient-matrix of dimension $p \times p_0$.
		
		\item Fit a local-logistic model based on reduced dataset $\{{\bf Z}_{\text{train}}, Y_{\text{train}}\}$ using the leave-one-out-cross-validation (LOOCV) criterion for bandwidth selection.
		\begin{enumerate}
			
			\item Let $h$ a bandwidth varying in a grid $\{h_1,h_2,\ldots, h_m\}$.
			
			\item Build the local-logistic model based on  $\{{\bf Z}_{\text{train}}, Y_{\text{train}}\}$ and estimate $\widetilde{R}_h({\bf z}_i)$ for all $i=1,2,\ldots, n_1$.
			
			\item For $i=1,2,\ldots, n_1$, let the leave-one-out ($loo$) dataset $\{{\bf Z}_{\text{train}}, Y_{\text{train}}\}^{(-i)}$, the dataset without the $i$th input. Let $\widetilde{R}_h^{(-i)}({\bf z}_{\text{train}})$ the fitted model using the $loo$ dataset and estimate $\widetilde{R}^{(-i)}_h({\bf z}_i)$ for all $i=1,2,\ldots, n_1$.
			
			\item Define the cross-validation score as 
			
			$$Q(h)=\sum_{i=1}^{n_1}  \widetilde{R}_h({\bf z}_i)^2-2\sum_{i=1}^{n_1} Y_i\widetilde{R}_h^{(-i)}({\bf z}_{\text{train}}).$$
			
			\item Define $h_{CV}=\arg \min_h \ Q(h)$.
			
		\end{enumerate}
		
		\item Let ${\bf x}_{0,\text{train}}$, a particular configuration of states of the components of the system in the train set. Let ${\bf z}_{0,\text{train}}=\left({\bf x}_{0,\text{train}}-(\mu_1, \mu_2, \ldots, \mu_p)\right){\boldsymbol \Sigma}_0^{-1}{\boldsymbol \Gamma_{\text{train}}}$, and let us build a local-logistic model. Then, for  $ {\bf z}_\text{train}$ such that $\|{\bf z}_\text{train}- {\bf z}_{0,\text{train}}\| < h_{CV}$, the fitted model is given by
		
		\[
		\widetilde{R}_{CV}({\bf z}_\text{train})= \frac{e^{ (1,({\bf z}_\text{train}-{\bf z}_{0,\text{train}})^t){\bf \widehat{b}}_\text{train}}}{1+e^{ (1,({\bf z}_\text{train}-{\bf z}_{0,\text{train}})^t){\bf \widehat{b}}_\text{train}}},
		\]
		
		with ${\bf \widehat{b}}_\text{train} =(\widehat{b}_0,\widehat{b}_1,\ldots, \widehat{b}_{p_0})^t$ vector of coefficients obtained using $h_{CV}$. 
		
		\item Let ${\bf x}_{0,\text{train}}$ such that $\|({\bf x}_\text{train}- {\bf x}_{0,\text{train}}) {\boldsymbol \Sigma}_0^{-1} {\boldsymbol \Gamma}_{\text{train}}\| < h_{CV}$, then the local-logistic model in the state space of components is estimated by
		
		\[
		\widehat{R}^*_{CV}({\bf x}_\text{train})= \frac{e^{(1,({\bf x}_\text{train}-{\bf x}_{0,\text{train}})^t){\widehat{\beta}_{\text{train}}}}}{1+e^{(1,({\bf x}_\text{train}-{\bf x}_{0,\text{train}})^t){\widehat{\beta}_{\text{train}} }}},
		\]
		
		where the vector of estimated coefficients ${\widehat{\beta}_{\text{train}}}$ is
		
		\begin{eqnarray*}
			\widehat{\beta}_0&=&\widehat{b}_0,\\
			\widehat{\beta}_j&=&\sigma_j^{-1}\boldsymbol{\Gamma}_{j\cdot} \widehat{\bf b}_{-0}, \hspace{0.5cm} j=1,2,\ldots, p;
		\end{eqnarray*}
		
		where we denote  $\boldsymbol{\Gamma}_{j\cdot}$ the $j$th row of matrix $\boldsymbol{\Gamma}_{\text{train}}$, and $\widehat{\bf b}_{-0}=(b_1,\ldots, b_{p_0})^t$.
		
		\item Define $\widehat{R}_{CV}^*({\bf x}_{i,\text{train}})=\frac{e^{b_0}}{1+e^{b_0}}$, for $i=1,2, \ldots, n_1$. Use the isotonization algorithm described in \cite{Gamiz2021} to isotonize these estimated values and then obtain the estimated reliability for the $i$th configuration of states of components, that is $\widehat{R}_{CV}({\bf x}_{i,\text{train}})$, $i=1,2,\ldots, n_1$.
		
		\item Labeled train data analyzed using the ROC curve to compute the ($AUC$) as in (\ref{eq:AUC}), and evaluate the model. 
	\end{enumerate}
	\item Working with test dataset:
	\begin{enumerate}
		\item Center the columns of matrix ${\bf X}_{\text{test}}$ that is, for each $j=1,2,\ldots, p$, define $U_j=(X_j-\mu_j)/\sigma_j$. Let $ {\boldsymbol \Sigma}_0$ a diagonal matrix with $j$th element $ \sigma_j$, and ${\bf M}_{\text{test}}={\bf 1}_p (\mu_1, \mu_2, \ldots, \mu_p)$, with ${\bf 1}_p$ a unitary column vector of size $p$. Let us denote ${\bf U}_{\text{test}}= \left({\bf X}_{\text{test}}- {\bf M}_{\text{test}}\right) {\boldsymbol \Sigma}_0^{-1}$, then ${\bf U}_{\text{test}}$ is a matrix with dimension $n_2 \times p$.
		
		\item Transform the data to a reduced set denoted $\{{\bf Z}_{\text{test}}, {Y_{\text{test}}}\}_{ n_2 \times p_0}$, with $p_0 < p$, and ${\bf Z}_{\text{test}} = {\bf U}_{\text{test}} \boldsymbol{\Gamma}_{\text{train}}$, where $\boldsymbol{\Gamma}_{\text{train}}$ is $p \times p_0$.

		\item Let ${\bf x}_{0,\text{test}}$, a particular configuration of states of the components of the system in the test dataset. Let ${\bf z}_{0,\text{test}}=\left({\bf x}_{0,\text{test}}-(\mu_1, \mu_2, \ldots, \mu_p)\right){\boldsymbol \Sigma}_0^{-1}{\boldsymbol \Gamma}_{\text{train}}$, and let us build a local-logistic model. Then, for  $ {\bf z}_{\text{test}}$ such that $\|{\bf z}_{\text{test}}- {\bf z}_{0,\text{test}}\| < h_{CV}$, the fitted model is given by
		
		\[
		\widetilde{R}_{CV}({\bf z}_{\text{test}})= \frac{e^{ (1,({\bf z}_{\text{test}}-{\bf z}_{0,\text{test}})^t){\bf \widehat{b}}_{\text{test}}}}{1+e^{ (1,({\bf z}_{\text{test}}-{\bf z}_{0,\text{test}})^t){\bf \widehat{b}}_{\text{test}}}},
		\]
		
		with ${\bf \widehat{b}}_{\text{test}} =(\widehat{b}_0,\widehat{b}_1,\ldots, \widehat{b}_{p_0})^t$ vector of coefficients obtained using $h_{CV}$. 
		
		\item Let ${\bf x}_{0,\text{test}}$ such that $\|({\bf x_{\text{test}}- {\bf x}_{0,\text{test}}) {\boldsymbol \Sigma}_0^{-1} {\boldsymbol \Gamma}_{\text{train}}}\| < h_{CV}$, then the local-logistic model in the state space of components is estimated by
		
		\[
		\widehat{R}^*_{CV}({\bf x}_{\text{test}})= \frac{e^{(1,({\bf x}_{\text{test}}-{\bf x}_{0,\text{test}})^t){\widehat{\beta}_{\text{test}}}}}{1+e^{(1,({\bf x}_{\text{test}}-{\bf x}_{0,\text{test}})^t){\widehat{\beta}_{\text{test}} }}},
		\]
		
		where the vector of estimated coefficients ${\widehat{\beta}_{\text{test}}}$ is
		
		\begin{eqnarray*}
			\widehat{\beta}_0&=&\widehat{b}_0,\\
			\widehat{\beta}_j&=&\sigma_j^{-1}\boldsymbol{\Gamma}_{j\cdot} \widehat{\bf b}_{-0}, \hspace{0.5cm} j=1,2,\ldots, p;
		\end{eqnarray*}
		
		where we denote  $\boldsymbol{\Gamma}_{j\cdot}$ the $j$th row of matrix $\boldsymbol{\Gamma_{\textbf{train}}}$, and $\widehat{\bf b}_{-0}=(b_1,\ldots, b_{p_0})^t$.
		
		\item Define $\widehat{R}_{CV}^*({\bf x}_{i,\text{test}})=\frac{e^{b_0}}{1+e^{b_0}}$, for $i=1,2, \ldots, n_2$. Use the isotonization algorithm described in \cite{Gamiz2021} to isotonize these estimated values and then obtain the estimated reliability for the $i$th configuration of states of components, that is $\widehat{R}_{CV}({\bf x}_{i,\text{test}})$, $i=1,2,\ldots, n_2$.
		
		\item Labeled train data analyzed using the ROC curve to compute the ($AUC$) as in (\ref{eq:AUC}), and evaluate the model. 
	\end{enumerate}
	
\end{enumerate}

\section{Numerical Results}\label{Sec:results}
In this section, we present a simulation study of several systems, along with an application to a real data set. These analyses demonstrate the strong performance of the FA-LR-IS algorithm compared to the supervised learning methods mentioned above.

\subsection{Simulations}
\begin{figure}[ht!]
	\centering
	\includegraphics[width=0.5\textwidth]{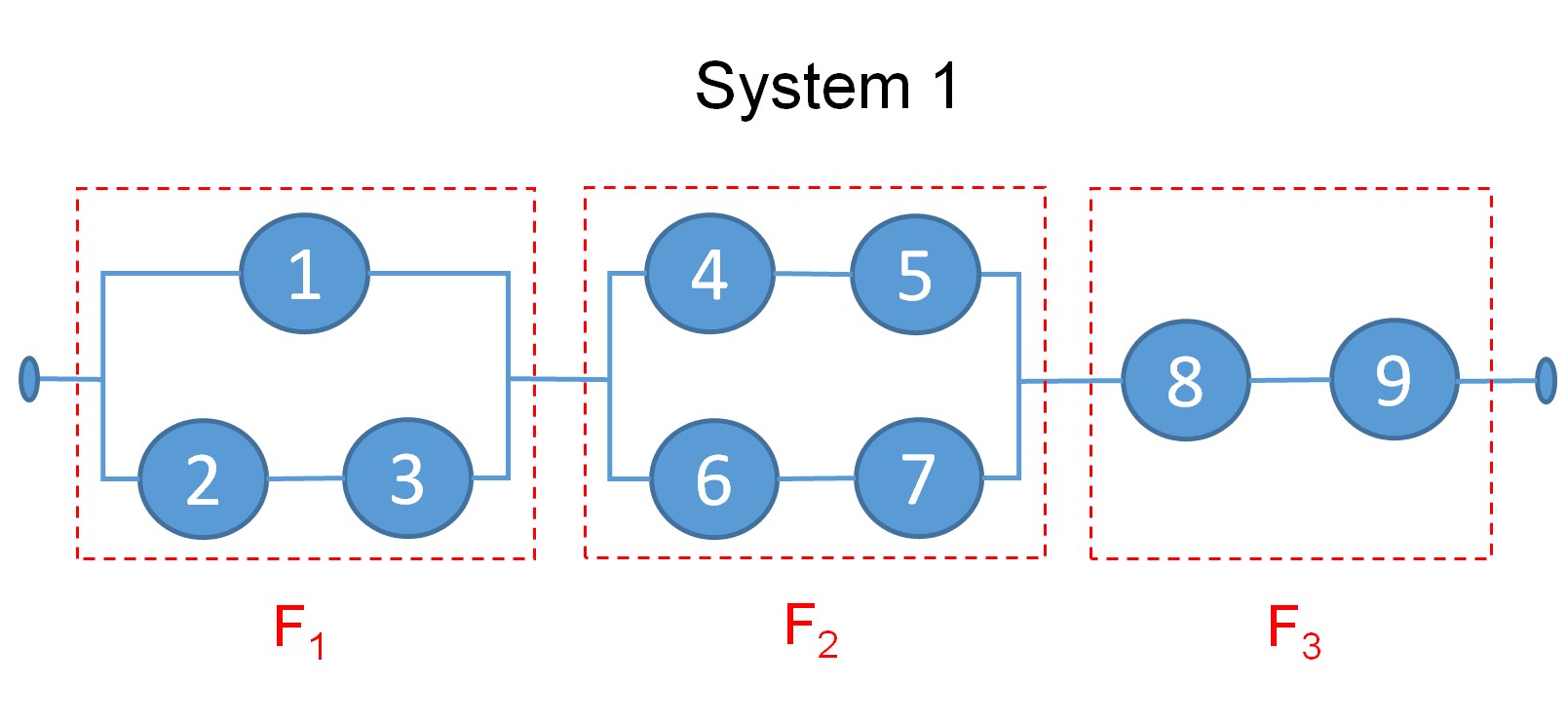} 
	\includegraphics[width=0.5\textwidth]{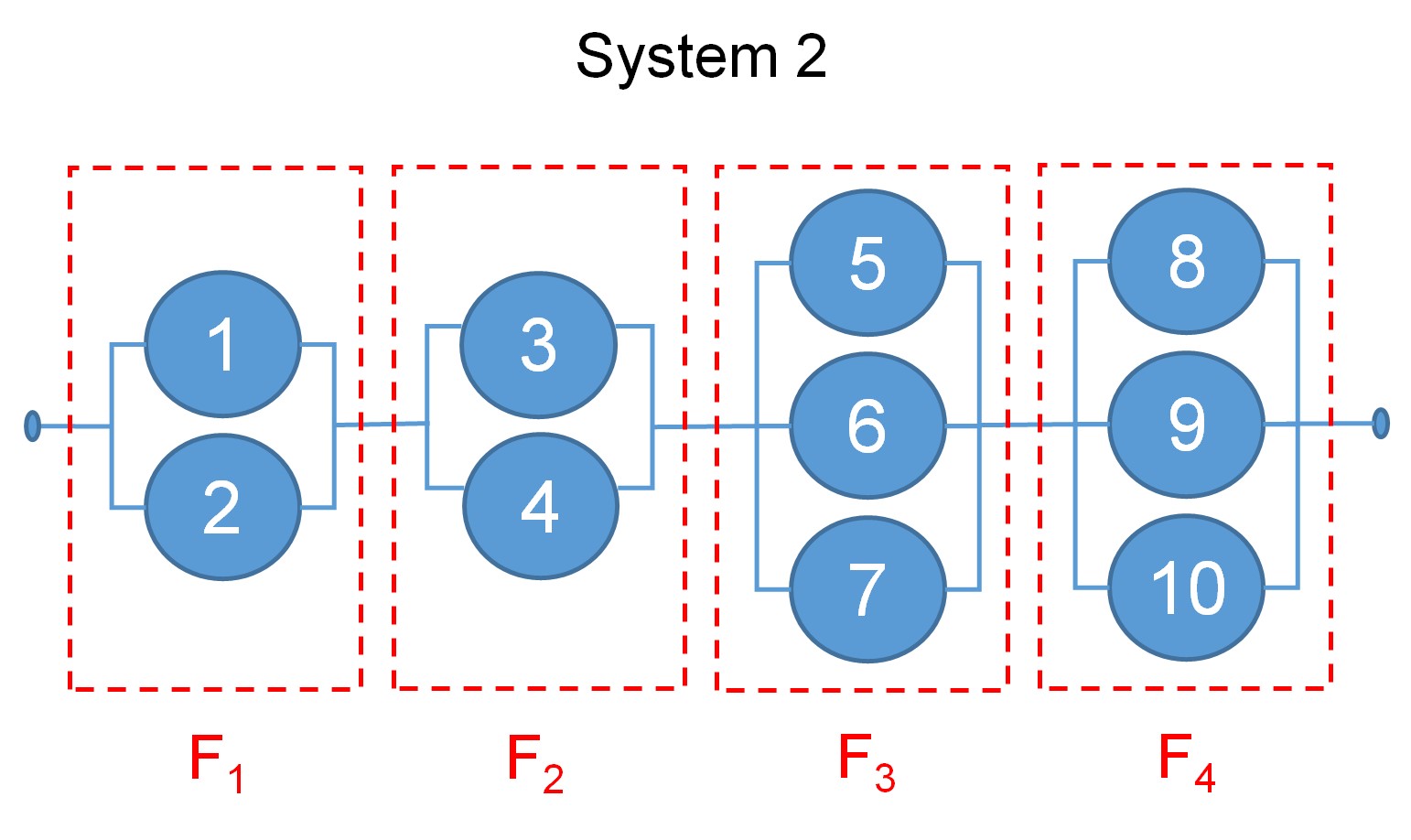}
	\includegraphics[width=0.6\textwidth]{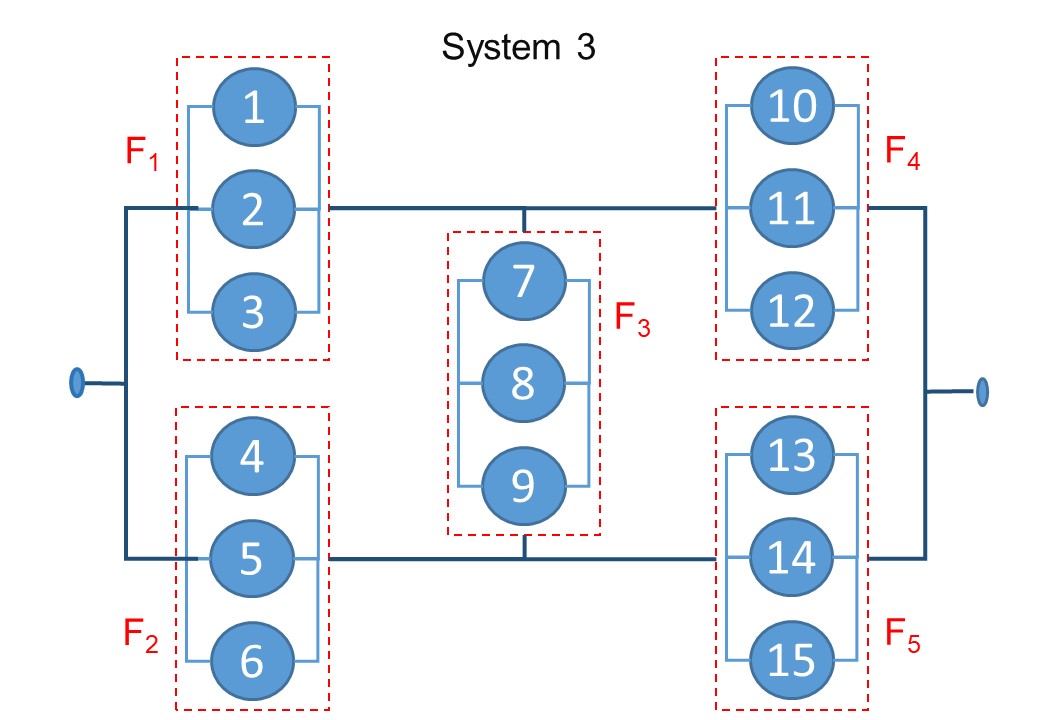}
	\includegraphics[width=0.6\textwidth]{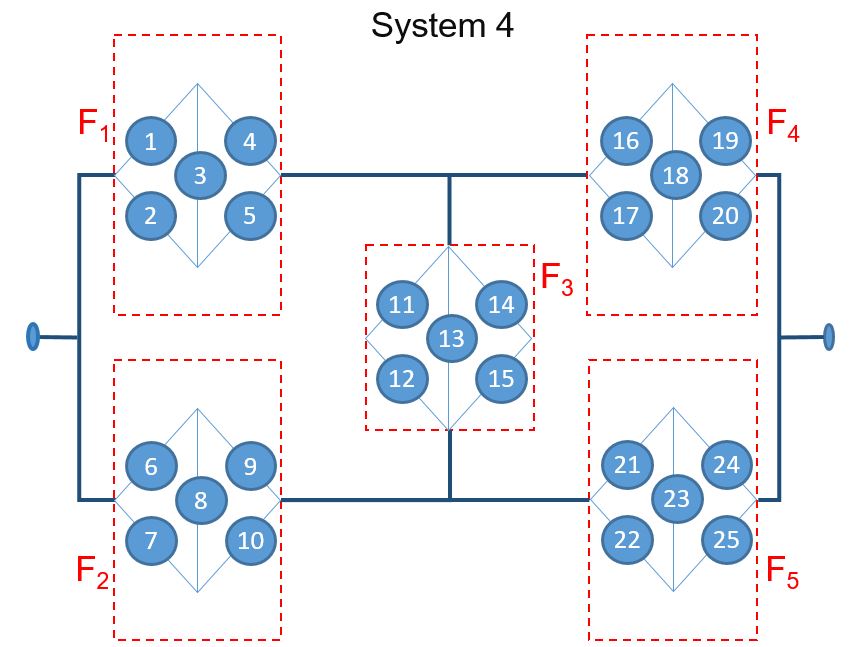}
	\caption{Reliability Block Diagrams for the simulated systems.}
	\label{fig:models}
\end{figure}

To evaluate our procedure, we conducted a simulation study with systems based on different configurations. Many scientific fields involve static or dynamic systems composed of multiple components, which can be grouped into distinct, interacting blocks. We assume that the system’s structural logic can be represented using a block diagram. Figure \ref{fig:models} presents a graphical representation of the four cases analyzed in this section.  

The data for each case have been generated as follows. We simulate 500 samples with size $n=125$. Let $p$ denote the number of components of the system, then $p=9, 10, 15, 25$  respectively, in our examples. The data consists of a matrix with $p+1$ columns. The first $p$ columns report the state of the components $X_{1},X_{2},\ldots,X_{p}$ where each $X_k \in [0,1]$, $k=1,\ldots,p$; while the $p+1$ column refers to the state of system $Y$, which takes value 1 if the system is in operation and 0 otherwise. Then the components are combined in blocks, denoted as $F_j$, for $j = 1, 2, 3$. for example, in System 1. We assume that $cor(X_k,X_{k'})=0.9$ if components $k$ and $k'$ are disposed in the same block $F$; and  $cor(X_k,X_{k'})=0$ otherwise. This point is considered in Figure \ref{fig:models}, where the blocks of dependent components have been highlighted using dashed lines.

The state of the system is simulated considering a latent variable that is not directly observed and is assumed to be $Y\rightarrow N(\Phi(\textbf{X}),\sigma)$, with  $\textbf{X}=(X_{1},X_{2},\ldots,X_{p})$ being a particular configuration of the state vector. We fix $\sigma=0.2$ and finally, the information about the state of system $Y$ is simulated from a Binomial distribution with event probability given by $R(\textbf{x})=P(Y>y_{0})$, with $y_{0}=0.5$.

The structure function of each model is given in the following.
\begin{itemize}

	\item  {\bf  System 1.} We consider a series-parallel system with $p=9$ components, as shown in Figure \ref{fig:models} (top plot). The system is composed of three blocks connected in series. The first two blocks are arranged in parallel and have three and four components, respectively. The third block consists of two components in series. In this context, the structure function of the system is given by the following expression
	$$
	\phi(\textbf{x})=\min(\max(x_{1},\min(x_{2},x_{3})),\max(\min(x_{4},x_{5}),  \min(x_{6},x_{7})),\min(x_{8},x_{9})),
	$$
	where $x_{j}$ denotes the state of the $j$th component, $j=1,2,\ldots,9$.

	\item {\bf System 2.} We consider a series-parallel combination system with $p=10$ components, as displayed in Figure \ref{fig:models} (second plot). The system is composed of four parallel blocks connected in series. The first two blocks have two components each and the next two blocks have three components each. In this case, the structure function of the system is given by the following expression
	$$
	\phi(\textbf{x})=\min(\max(x_{1},x_{2}),\max(x_{3},x_{4}),
	\max(x_{5},x_{6},x_{7}),\max(x_{8},x_{9},x_{10})),
	$$
	
	where $x_{j}$ denotes the state of the $j$th component, $j=1,2,\ldots,10$.

	\item {\bf System 3.} We consider a bridge structure with $p=15$ components as displayed in the bottom plot of Figure \ref{fig:models} (third plot). A simple bridge structure has been modified introducing redundancy \cite{Sahoo14}. That is, each node has been replaced by a block consisting of three units connected in parallel.
	\begin{equation*}
		\phi(\textbf{x})=\max(\min({\bf x}_1,{\bf x}_4),\min({\bf x}_1,{\bf x}_3,{\bf x}_5),\min({\bf x}_2,{\bf x}_3,{\bf x}_4),\min({\bf x}_2,{\bf x}_5))
	\end{equation*}
	where ${\bf x}_{k}=\max(x_{3k-2},x_{3k-1},x_{3k})$ , for $k =1, \ldots, 5$, and $x_j$ denoting the state of the $j$th component, $j=1,2,\ldots,15$.
	
	\item {\bf System 4.} We consider a bridge structure with $p=25$ components as displayed in the bottom plot of Figure \ref{fig:models} (bottom plot). A simple bridge structure has been modified introducing redundancy. That is, each node has been replaced by a block consisting of a bridge structure with five components.
	\begin{equation*}
		\phi(\textbf{x})=\max(\min({\bf x}_1,{\bf x}_4),\min({\bf x}_1,{\bf x}_3,{\bf x}_5),\min({\bf x}_2,{\bf x}_3,{\bf x}_4),\min({\bf x}_2,{\bf x}_5))
	\end{equation*}
	where 
	\begin{eqnarray*}
		&& {\bf x}_{j}=\max(\min(x_{5j-4},x_{5j-1}),\min(x_{5j-4},x_{5j-2},x_{5j}), \\
		&& \hspace{2cm} \min(x_{5j-3},x_{5j-2},x_{5j-1}),\min(x_{5j-3},x_{5j})),
	\end{eqnarray*}
	for $j =1, \ldots, 5$, and $x_k$ denoting the state of the $k$th component, $k=1,2,\ldots,25$.
\end{itemize}

Each sample $\chi = \{(x_i,y_i), i=1,\ldots,n\}$ is split into a training set and a test set, as illustrated in Figure \ref{fig:tt}. We then have $\chi = \{\chi_1 \cup \chi_2\}$, with the corresponding set of indices $I=\{1,2,\ldots,125\},$ similarly divided into $I=\{I_1 \cup I_2\}$, where $I_1$ represents the indices of the training set and $I_2$ represents the indices of the test set. Accordingly:
\begin{eqnarray*}
	\text{car}(I_1)  = 0.8 \times n = n_1 \\
	\text{car}(I_2)  = n - n_1 = n_2
\end{eqnarray*}

The results presented below were obtained using the estimated reliability for each system ($S$=1,2,3,4) with a sample size of $n=125$, ensuring that the sample split remains consistent to facilitate reproducibility across methods. This guarantees that all algorithms are applied to the same datasets.

\begin{enumerate}
	
	\item The FA-LR-IS algorithm begins by reducing the number of features using a factor analysis (FA) approach. This step identifies optimal factors representing the system.  To determine the appropriate number of factors (or blocks) required for dimensionality reduction, we employed the testing procedure implemented in the {\it factanal} function in the R software, as described in \cite{Gamizetal2023}. The goal is to maintain an effective reduction in dimensionality without significant loss of information. Additionally, bandwidth, another critical parameter for this method, was selected using cross-validation techniques, as explained in Section \ref{sec:algorithm}. 
	
	\item For the Artificial Neural Network (ANN) model, we employed a feedforward neural network with three layers, aimed at classifying the system state (0: failure, 1: operative). The ANN architecture includes 50 input neurons, two hidden layers with 15 and 80 neurons respectively (both using ReLU activation), and a final output neuron with sigmoid activation for binary classification. The network was optimized using the ADAM algorithm with the binary cross-entropy loss function. The ADAM algorithm, which combines AdaGrad and RMSprop, was chosen due to its adaptability, ease of implementation, and computational efficiency, \cite{Kingma2017}. We ran the model for 125 epochs with a batch size of 64. The architecture (layers and neurons) was selected by evaluating various brute-force combinations.
	
	\item We used a $K$-Nearest Neighbors (KNN) classifier with the number of neighbors set to 20, a value determined to be optimal through brute-force search. The distance metric used was Minkowski distance. 
	
	\item For RF we used a Random Forest Classifier with 100 estimators, which were chosen through trial and error to balance performance without overfitting.
\end{enumerate}

Below, we present a series of tables and graphs comparing the FA-LR-IS algorithm with other ML methods.

\begin{figure}[ht!]
		\centering
		\includegraphics[width=0.7\textwidth]{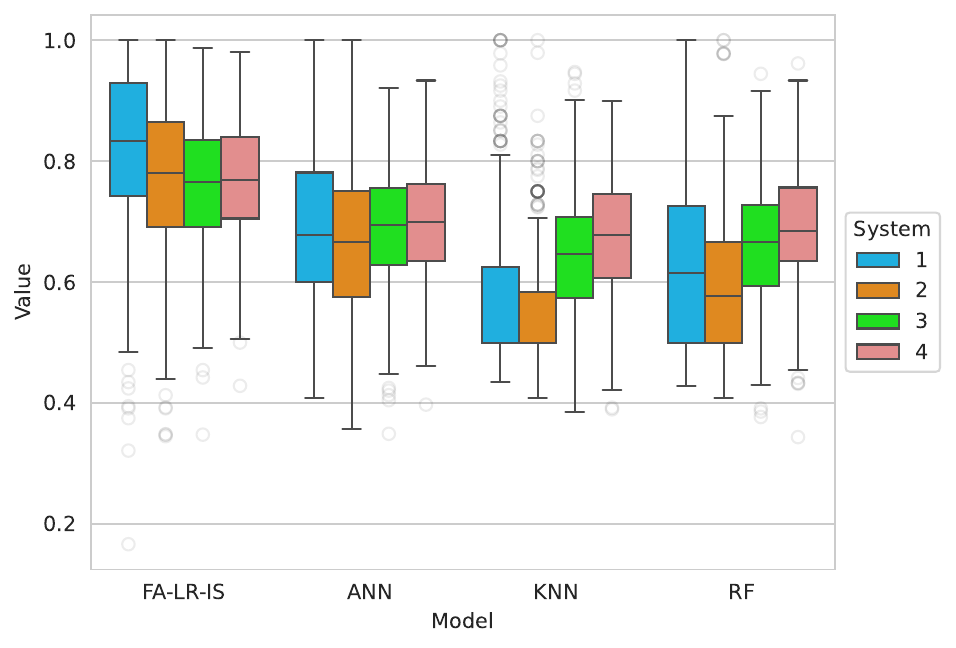}
	\caption{Boxplot comparing $AUC$ between models.\label{AUC2}}
\end{figure}

Figure \ref{AUC2} displays the results of goodness of fit for each model, obtained by running the different algorithms. ($AUC$) was used to assess the goodness of fit. Each system is color-coded for clarity. 
Table  \ref{tab:AUC} shows the mean and standard deviation (SD) of the $AUC$ calculated over all repetitions in the experiment. The FA-LR-IS algorithm consistently outperforms the other methods across all system configurations, demonstrating superior discriminatory power as a classification method.
\begin{table}[H]
	\caption{Mean and SD of area under the ROC curve ($AUC$), defined in (\ref{eq:AUC}).\label{tab:AUC}}
		\begin{tabular}{lcccc}
		\hline
			\textbf{Models}    & \textbf{System 1}    & \textbf{System 2} & \textbf{System 3} & \textbf{System 4} \\
		\hline
			& Mean (SD)   &  Mean (SD) &  Mean (SD)  &  Mean (SD)  \\
		\hline
			FA-LR-IS & {\bf 0.8218} (0.1338) & {\bf 0.7708} (0.1273) & {\bf 0.7599} (0.1045) & {\bf 0.7674} (0.0992) \\
			ANN & 0.6902 (0.1423) & 0.6702 (0.1373) & 0.6907 (0.0970) & 0.6998 (0.0941) \\
			KNN & 0.5763 (0.1212) & 0.5510 (0.0905) & 0.6428 (0.0981) & 0.6754 (0.0929) \\
			RF & 0.6315 (0.1395) & 0.5928 (0.1110) & 0.6632 (0.1021) & 0.6894 (0.0933) \\
		\hline
		\end{tabular}
\end{table}

\begin{figure}[ht!]
		\centering
		\includegraphics[width=0.7\textwidth]{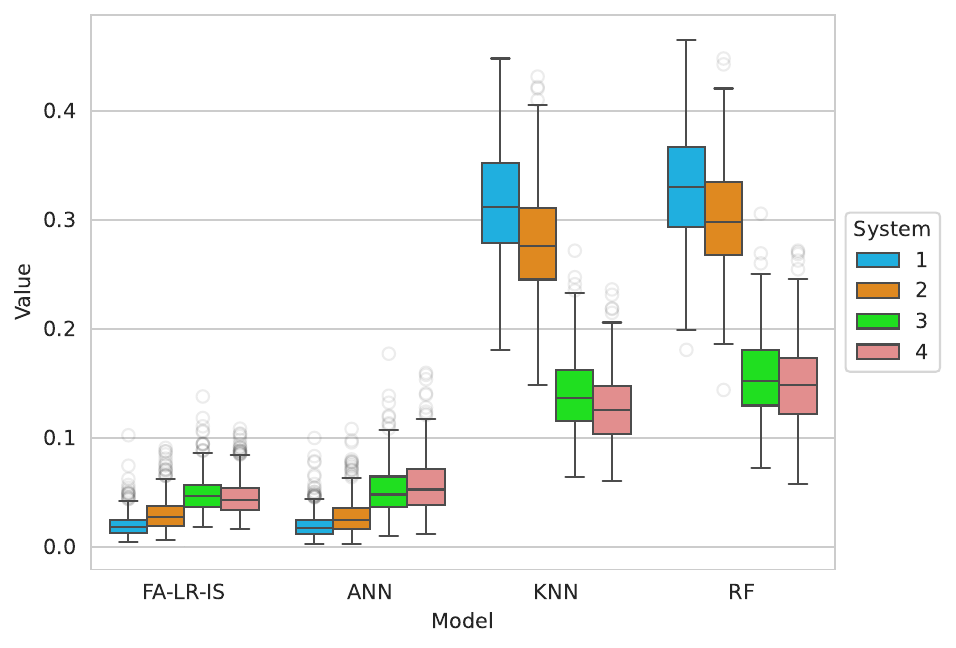}
	\caption{Boxplot comparing $MSE$ between models.}\label{ASE2}
\end{figure}

Figure \ref{ASE2} illustrates the accuracy of reliability estimation for each method. For each system, given a particular sample with a test dataset $\chi_2 =\{{\bf x}^m_{j},y^m_{j}; j=1,\ldots, n_{2}\}$, with $m =1,\ldots, M$, the mean square error $MSE_m$ was calculated using the formula:
\begin{equation}
	MSE_{S,m}^{\bullet}= \frac{1}{n_{2}} \sum_{j=1}^{n_{2}} \left(\widehat{R}_{S,j}^{\bullet,m}-R_S({\bf x}_{j}^m)\right)^2,
	\label{eq:MSE}
\end{equation}
where $R_S({\bf x}_j^m)$ represents the true reliability function for the structure $S$ and $\widehat{R}_{S,j}^{\bullet,m}$ denotes the reliability estimated using the corresponding method. Table \ref{tab:ASE} presents the average values and standard deviations (SD) of $MSE_{S,m}^{\bullet}$ across $M$ repetitions. As seen in Figure \ref{ASE2}, the FA-LR-IS and ANN perform similarly concerning MSE for the four systems considered.
However, Table \ref{tab:ASE} shows that FA-LR-IS outperforms all other methods for Systems 3 and 4, achieving the lowest error, while ANN performs comparably for Systems 1 and 2.

\begin{table}[H]
	\centering
	\caption{Mean and SD of $MSE$ for each system and model defined in (\ref{eq:MSE}).}\label{tab:ASE}
	
		\begin{tabular}{lcccc}
		\hline
			\textbf{Models}    & \textbf{System 1}    & \textbf{System 2} & \textbf{System 3} & \textbf{System 4}\\
		\hline
			& Mean (SD)   &  Mean (SD) &  Mean (SD)  &  Mean (SD)  \\
			\hline
			FA-LR-IS & 0.0202 (0.0107) & 0.0298 (0.0139) &  {\bf 0.0487} (0.0165) & {\bf 0.0456} (0.0163) \\
			ANN & {\bf 0.0199} (0.0122) & {\bf 0.0278} (0.0159) & 0.0515 (0.0214) & 0.0565 (0.0245) \\
			KNN    & 0.3150 (0.0521) & 0.2787 (0.0494) & 0.1406 (0.0340) & 0.1273 (0.0319) \\
			RF & 0.3307 (0.0519) & 0.3009 (0.0490) & 0.1568 (0.0379) & 0.1503 (0.0381) \\
			
		\hline
		\end{tabular}
\end{table}

\begin{figure}[ht!]
		\centering
		\includegraphics[width=0.7\textwidth]{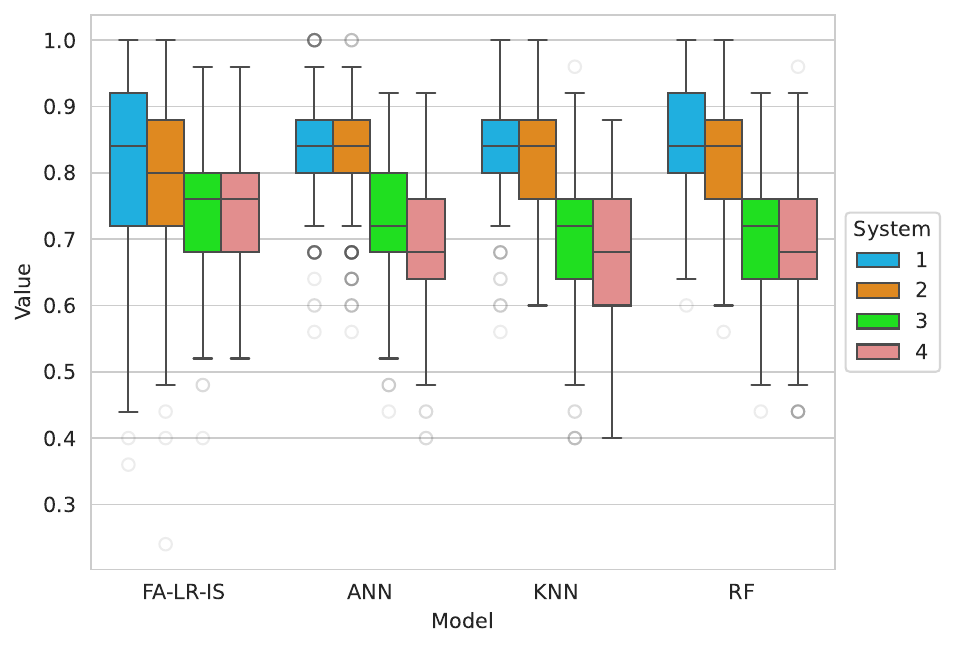}
	\caption{Boxplot comparing predictive capacity between models.\label{Accuracy2}}
\end{figure}

In figure \ref{Accuracy2}, the accuracy of the estimator, measured in terms of predictive capacity as defined in Section \ref{sec:metrics}, is shown. Table \ref{tab:Accuracy} provides the mean and SD for these results across repetitions. FA-LR-IS outperforms the other methods for Systems 3 and 4, while ANN performs best for System 1 and KNN for System 2.
\begin{table}[H]
	\centering
	\caption{Accuracy, measured as indicated in Section \ref{sec:metrics}.} \label{tab:Accuracy}
		\begin{tabular}{lcccc}
		\hline
			\textbf{Models}    & \textbf{System 1}    & \textbf{System 2} & \textbf{System 3} & \textbf{System 4}\\
		\hline
			& Mean (SD)   &  Mean (SD) &  Mean (SD)  &  Mean (SD)  \\
		\hline
			FA-LR-IS & 0.8166 (0.1214) & 0.7786 (0.1166) & {\bf 0.7523} (0.0891) & {\bf 0.7570} (0.0845) \\
			ANN    & {\bf 0.8482} (0.0752) & 0.8259 (0.0764) & 0.7245 (0.0854) & 0.6990 (0.0941) \\
			KNN & 0.8440 (0.0746) & 0.8250 (0.0763) & 0.7010 (0.0902) & 0.6743 (0.0930) \\
			RF     & 0.8482 (0.0730) & {\bf 0.8263} (0.0763) & 0.7068 (0.0919) & 0.6880 (0.0927) \\
			
		\hline
		\end{tabular}
\end{table} 

Finally, given that FA-LR-IS and ANN were the most competitive methods, we conducted direct comparisons bootstrap between them. For each measure ($AUC$, $MSE$ or Accuracy) the following procedure was implemented:

\noindent \textbf{Steps}
\begin{enumerate}
	\item[{\bf 1.}] For the 500 sample realizations of the measure obtained for each method (FA-LR-IS and ANN), calculate the difference between the means.
	\item[{\bf 2.}] Combine all results from both methods and draw two samples with replacements from this combined data, each of size 500. Calculate the means of these samples and then compute the difference between them.
	\item[{\bf 3.}] Repeat Steps 1 and 2 a total of 10,000 times to obtain an asymptotic bootstrap distribution for the difference in means.
	\item[{\bf 4.}] Calculate the $p$-value as the proportion of bootstrap differences that are less than or equal to the absolute value of the observed difference in means from the initial 500 observations. 
\end{enumerate}
Table \ref{tab:pvalue} shows the $p$-values for hypothesis tests assessing the equality of the measurements obtained with each method. For $AUC$ and $MSE$, FA-LR-IS significantly outperforms ANN ($p$-value < 0.0001). Although ANN achieved better results in accuracy for Systems 1 and 2, the differences were not statistically significant. In contrast, FA-LR-IS significantly outperformed ANN in Systems 3 and 4. Notably, when FA-LR-IS was superior, the difference was statistically significant, whereas the advantage of ANN was not significant in the cases where it performed better.

\begin{table}[H]
	\centering
	\caption{Comparing the $p$-value for each of the metrics for FA-LR-IS and ANN.}\label{tab:pvalue}
		\begin{tabular}{lcccc}
			\hline
			\textbf{Metrics}    & \textbf{System 1}    & \textbf{System 2} & \textbf{System 3} & \textbf{System 4} \\
		\hline
			$AUC$ & <0.0001 & <0.0001 & <0.0001 & <0.0001 \\
			$MSE$ & <0.0001 & <0.0001 & <0.0001 & <0.0001 \\
			Accuracy & 0.5986 & 0.0304 & 0.0198 & <0.0001 \\
		\hline
		\end{tabular}
\end{table}

\subsection{Real dataset: Water pump sensor data}\label{sec:realdata}
We analyze a dataset concerning an industrial structure, in particular, we have performance measurements of a water pump installed in a small area. The dataset was sourced from the data platform \url{www.kaggle.com}, and a statistical analysis of this data is presented in \cite{Alargasamy2021}. The sample information was collected by a set of sensors monitoring various components of the water pump over time. Specifically, 50 sensors measured parameters such as temperature, pressure, vibration, load capacity, volume, and flow density, among others, every minute from April 1st, 2018 to August 31th, 2018. In total, there are 220.320 observations. 

In our analysis, we did not use all available records, but instead sampled data at one-hour intervals, resulting in a sample size of $n=3672$. Limited information is available regarding the behavior of the sensors, but according to the study of one of the experts who have analyzed the data (available on the data website), the intermediate group of sensors (sensor16-sensor36) corresponds to the performance of two impellers, while the first 14 sensors monitor aspects related to the engine.

We applied the same algorithms used in previous simulations. Specifically, we implemented the FA-RL-IS algorithm explained in Section \ref{sec:algorithm} following this procedure:

\begin{enumerate}
	\item We split the sample into training and test sets (Step 1.).
	\item Using the training data, we proceeded as follows:
	\begin{enumerate}
		\item {\bf Data normalization} (Step 2(a)). Since the scales of the sensor measurements vary, it was necessary to normalize the data to avoid the influence of variables with larger scales and ensure comparability.
		\item {\bf Factor analysis} (Step 2(b)): This step involved:
		\begin{enumerate}
			\item Examining possible correlations. The correlation matrix with all variables is shown in Figure \ref{cormat}. Some sensor groups, such as the intermediate group, show high positive correlations within the group.
			\begin{figure}[ht!]
					\centering
					\includegraphics[width=0.7\textwidth]{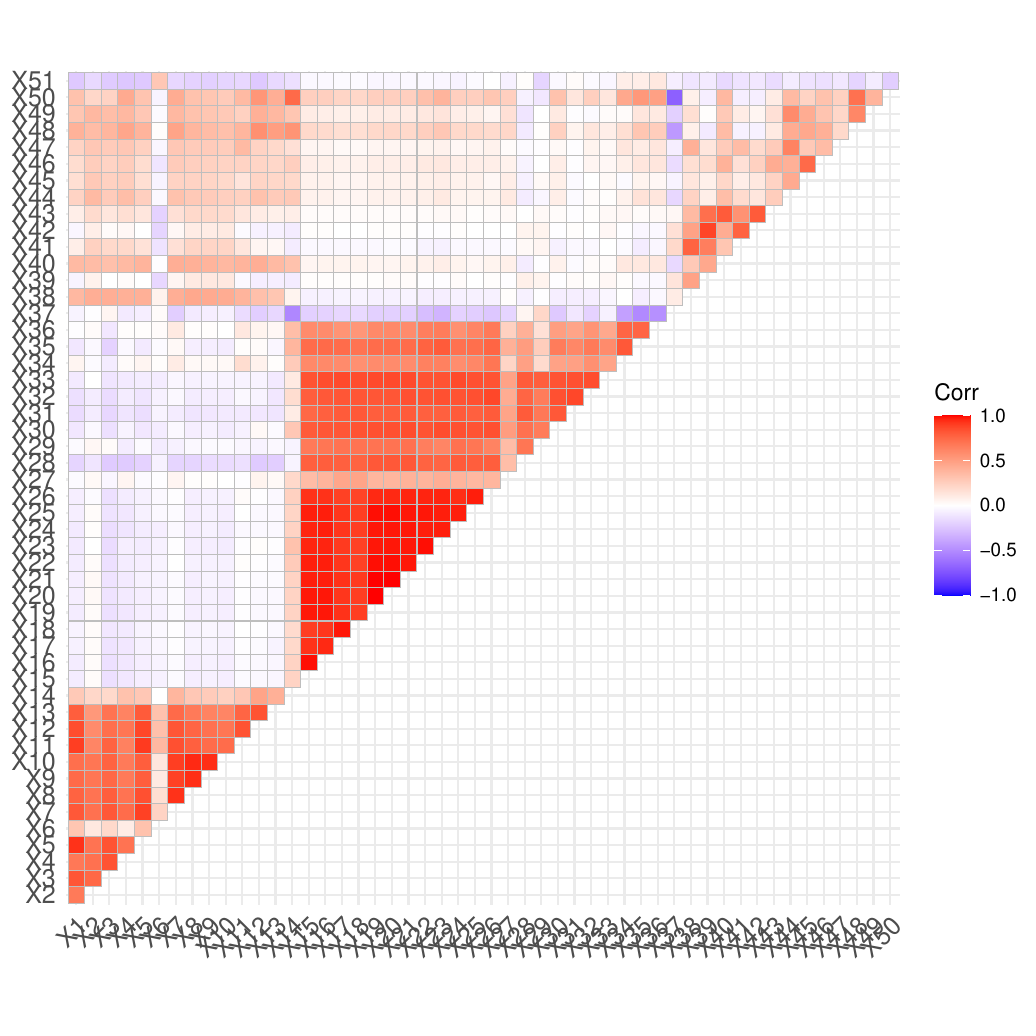}
				\caption{Correlation matrix.\label{cormat}}
			\end{figure}
			\item Determining the appropriate number of factors. Using the R package {\it psych} \cite{Revelle}, we determined the number of factors to extract, based on a {\it scree} plot (Figure \ref{screeplot}).
			\begin{figure}[ht!]
					\centering
					\hspace*{-10pt} 
					\includegraphics[width=0.7\textwidth]{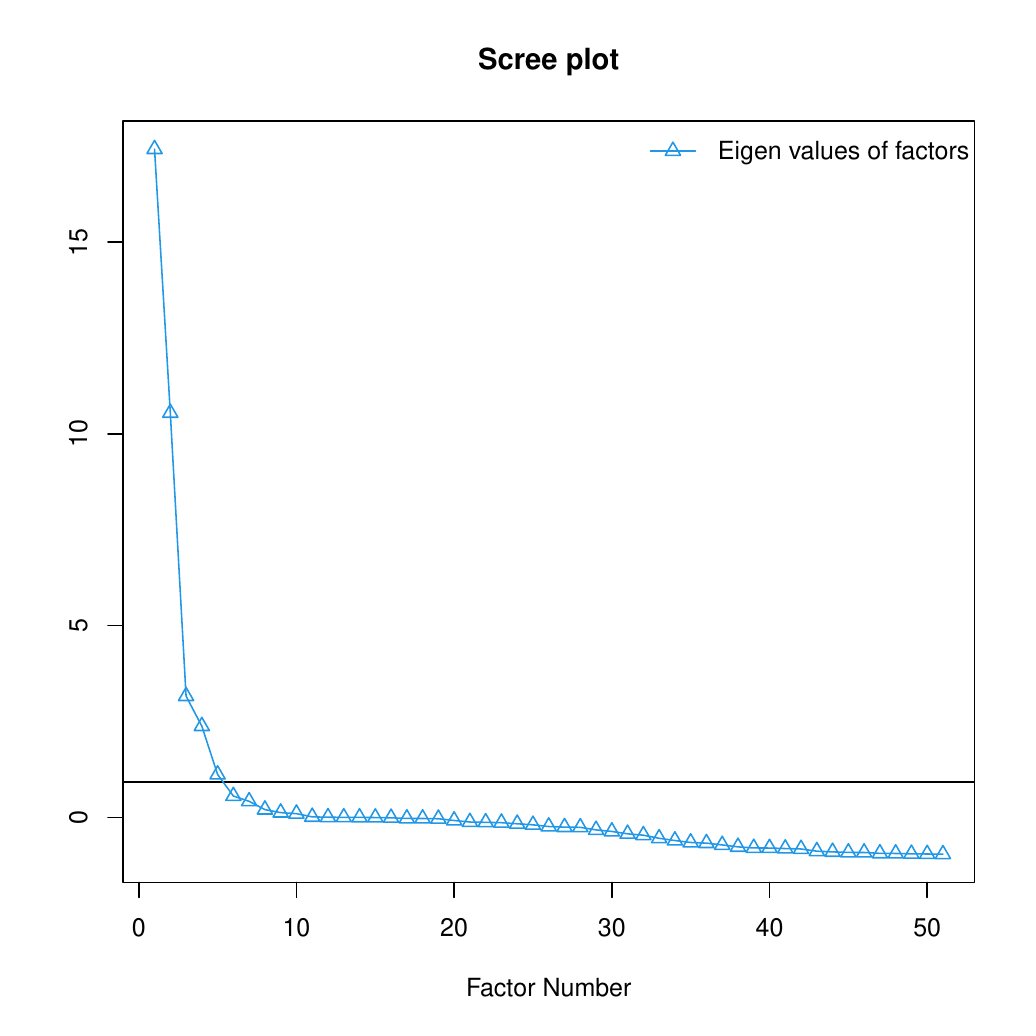}
				\caption{Scree plot. \label{screeplot}}
			\end{figure}
			\item Conducting the factor analysis. We used the {\it fa} function from the {\bf \it psych} package to perform an exploratory factor analysis of latent variables using maximum likelihood. The correlation matrix was decomposed into eigenvalues and eigenvectors, estimating the commonalities for each variable across the first five factors. Factor loadings and interfactor correlations were also obtained. 
		\end{enumerate}
		\item {\bf Local-logistic estimation}. We then fitted a local-logistic model in the space of the first five factors, $p_0=5$ (Step 2(d)), and back-transformed the results to the original feature space, $p=50$ (Step 2(e)). The bandwidth parameter was estimated through cross-validation (Step 2(c)). We now had a model that predicts the probability that the machine is functioning (reliability function) based on the sensor values. 
		\item {\bf Classification} (Step 2(g)). After estimating the probabilities with the logistic regression model, we translated these probabilities into classes or categories. For this, we used the classification obtained via the $\widehat{AUC}$, see (\ref{eq:AUC}).
		
	\end{enumerate}
	\item Working with the test dataset:
	\begin{enumerate}
		\item {\bf Normalize the data} (Step 3(a)). 
		\item {\bf Transform the data to a reduced set} (Step 3(b)).
		\item {\bf Local-logistic estimation}. Steps 3(c-d-e).
		
		We applied the same model used to predict the reliability function to the test dataset.
		\item {\bf Classification} (Step 3(f)). We classified the observations in the test set using the same method as with the training set.
		
	\end{enumerate}
	
	\item{Metrics}: All previous steps generated results to compute error metrics on the training dataset. We calculated these error metrics using the test dataset, fitting the local logistic model and applying the bandwidth obtained in Step 2(c) on the training data. The test set was classified following the procedure in Step 3(f).
	
\end{enumerate}

In addition to the proposed algorithm, we also executed the ANN, KNN, and RF algorithms using the same sample split and parameters as in the simulations. Table \ref{realestest} shows the error metrics calculated on the test dataset. We observe that the ANN algorithm yields the highest sensitivity, accuracy, and F1-score, while KNN and RF show the highest Specificity and TPV. The proposed algorithm shows high values comparable to the other ML algorithms, except for specificity, where it underperforms relative to ANN.
\begin{table}[H] 
	\centering
	\caption{Error metrics defined on Section \ref{sec:metrics} calculated on test dataset.\label{realestest}}
	\begin{tabular}{lcccc}
\hline 
		& \textbf{FA-LR}	& \textbf{ANN}	& \textbf{KNN} & \textbf{RF}\\
		\hline
		Sensitivity & 0.9845		&	{\bf 0.9971}		& 0.9852  & 0.9859\\
		Specificity & 0.9167		& 0.9583			& {\bf 0.9792} & {\bf 0.9792}\\
		Accuracy & 0.9809 & {\bf 0.9946} & 0.9904 & 0.9918\\
		TPV & 0.9941 & 0.9971& {\bf 0.9985} & {\bf 0.9985}\\
		F1-Score & 0.9898 & {\bf 0.9971} &  0.9949& 0.9956 \\
	\hline
	\end{tabular}
	
\end{table}

%

\section{Discussion}\label{Sec:disc}
One significant limitation of our study is the inability to train a deep learning model for comparison with our custom statistical model. While deep learning models typically outperform ANN, they require substantial datasets to learn effectively \cite{Bansal22}. Unfortunately, our current dataset is insufficient in size to support the training of such a model. Additionally, the computational resources necessary for training deep learning models are beyond our current capabilities.

An interesting avenue for future research could involve measuring the performance of FA-LR-IS against a deep learning model over time. We hypothesize that our statistical model will initially outperform deep learning models in scenarios where the system is new and data availability is limited, due to its lower dependency on large volumes of data. However, as the system matures and accumulates more data, the deep learning model may exhibit superior performance.

We aim to explore several questions in future work: At what point does this shift occur? How does it differ from system to system? Is the improvement worth the cost of deploying a deep learning model?

There are also models of ANNs that allow for feedback loops, known as recurrent neural networks (RNNs). While RNNs have been less influential than feedforward networks, partly because their learning algorithms are less powerful to date, they are still extremely interesting. RNNs more closely resemble the way our brains operate, and they may be capable of solving important problems that feedforward networks can only tackle with great difficulty.

\section{Conclusions}\label{Sec:conc}

In this paper, we introduce the FA-LR-IS algorithm, designed to build a statistical model capable of predicting system reliability for complex systems with numerous interdependent components. This method is entirely data-driven and does not rely on any parametric assumptions. We propose a procedure for dimensionality reduction, transforming a large set of inputs into an optimal subset, as discussed.

We compare the performance of the new algorithm in conducting reliability analyses of complex high-dimensional systems with various machine learning methods applicable in this context. Our algorithm performs as well as, or better than ANN in several cases.

To illustrate the method, we carried out a simulation study and applied it to a real dataset. In all instances, the FA-LR-IS algorithm yielded favorable results regarding block recognition and model accuracy.

Future work will focus on investigating the flexibility of the model in quantifying the localized effects of specific units on system performance, comparing these findings with the machine learning methodologies discussed in this paper, as well as other methods.



\end{document}